\documentclass{article} 
\usepackage{nips12submit_e,times}

\usepackage{graphicx}
\usepackage{psfrag,epsfig,subfigure}
\usepackage{amsmath,amssymb,amsfonts,amsbsy,amsthm}
\usepackage{latexsym}

\input{def.set}

\title{Regularized Discriminant Embedding for Visual Descriptor Learning}

\author{
Kye-Hyeon Kim,$^\mathrm{a}$ Rui Cai,$^\mathrm{b}$ Lei Zhang,$^\mathrm{b}$ Seungjin Choi$^\mathrm{a}$\thanks{The full version of this manuscript is currently under review in an international journal.} \\
$^\mathrm{a}$ Department of Computer Science, POSTECH, Pohang 790-784, Korea \\
$^\mathrm{b}$ Microsoft Research Asia, Beijing 100080, China \\
\texttt{fenrir@postech.ac.kr},~~ \texttt{\{ruicai, leizhang\}@microsoft.com}, \\ \texttt{seungjin@postech.ac.kr} \\
}

\nipsfinalcopy 

\begin{document}

\maketitle

\begin{abstract}
Images can vary according to changes in viewpoint, resolution, noise, and illumination.
In this paper, we aim to learn representations for an image, which are robust to wide changes in such environmental conditions, using training pairs of matching and non-matching local image patches that are collected under various environmental conditions.
We present a regularized discriminant analysis that emphasizes two challenging categories among the given training pairs: (1) matching, but far apart pairs and (2) non-matching, but close pairs in the original feature space (e.g., SIFT feature space).
Compared to existing work on metric learning and discriminant analysis, our method can better distinguish relevant images from irrelevant, but look-alike images.
\end{abstract}

\section{Introduction}
\label{sec:introduction}

In many computer vision problems, images are compared using their {\em local descriptors}.
A local descriptor is a feature vector, representing characteristics of an {\em interesting local part} in an image.
Scale-invariant feature transform (SIFT) \cite{LoweDG99iccv} is popularly used for extracting interesting parts and their local descriptors from an image.
Then comparing two images is done by aggregating pairs between each local descriptor in one image and its closest local descriptor in another image, whose pairwise distances are below some threshold.
The assumption behind this procedure is that
local descriptors corresponding to the same local part (``matching descriptors'') are usually close enough in the feature space, whereas local descriptors belonging to different local parts (``non-matching descriptors'') are far apart.

\begin{figure}[t]
\centering
\includegraphics[width=11.5cm]{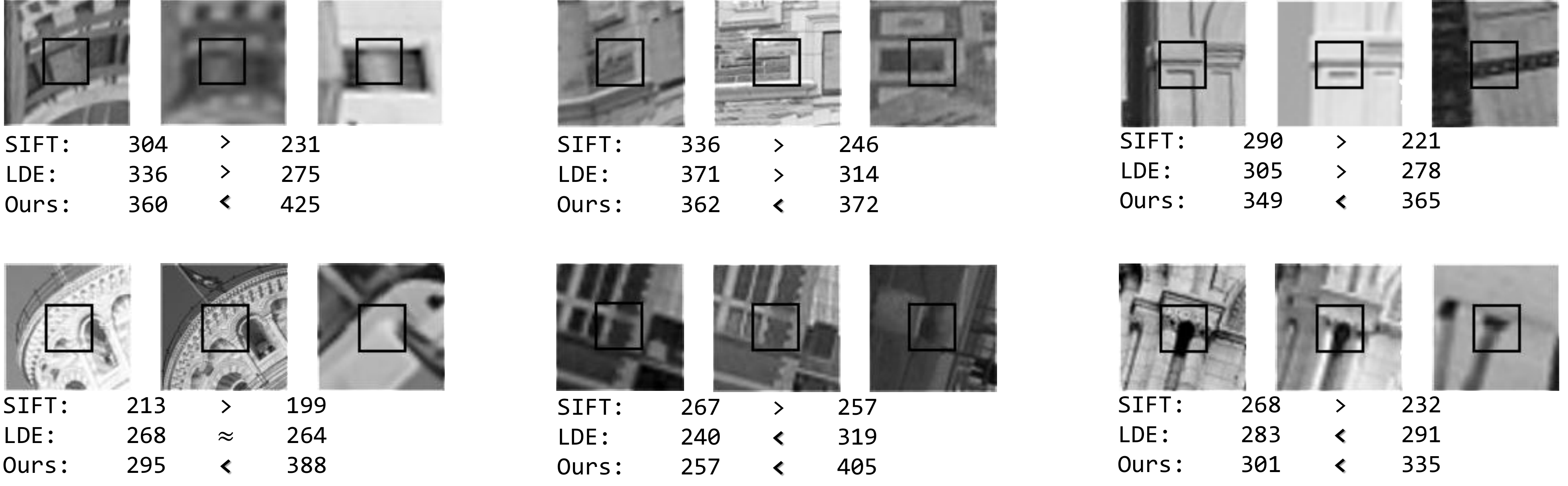}
\caption{Some examples where a local part (center in each triplet) is closer to a non-matching part (right) than a matching part (left) in terms of the Euclidean distances between their SIFT descriptors.
Using linear discriminant embedding (LDE) \cite{HuaG2007iccv}, non-matching pairs are still closer than matching pairs in the first three examples.
Compared to existing work on metric learning and discriminant analysis, our learning method focuses more on ``far but matching'' and ``close but non-matching'' training pairs, so that can distinguish look-alike irrelevant parts successfully.
}
\label{fig:example_relfar_irrnear}
\end{figure}

\begin{figure}[t]
\centerline{
\subfigure[15 closest SIFT pairs]{\includegraphics[width=4.1cm]{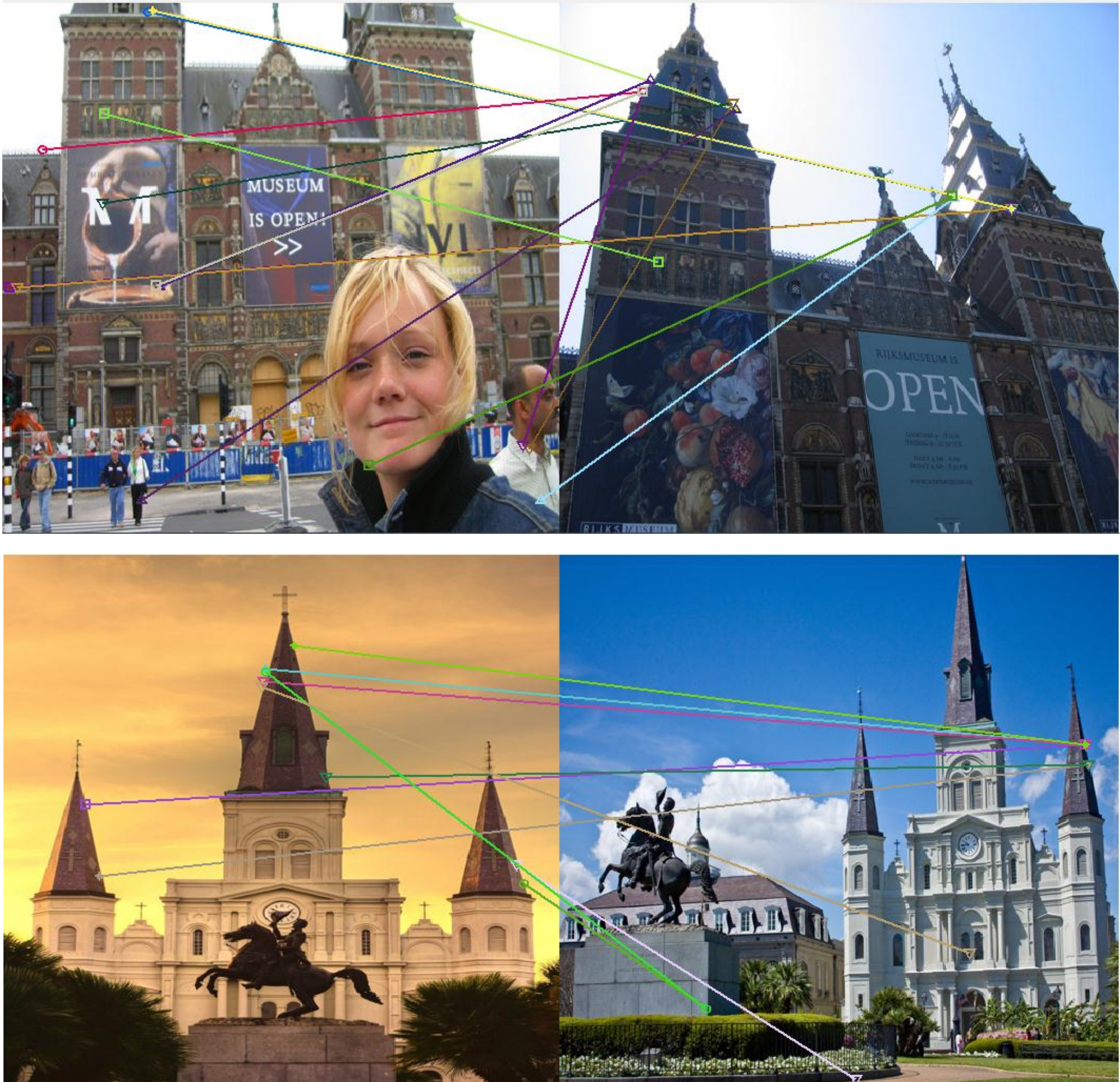}}
\hfil
\subfigure[15 closest RDE pairs]{\includegraphics[width=4.1cm]{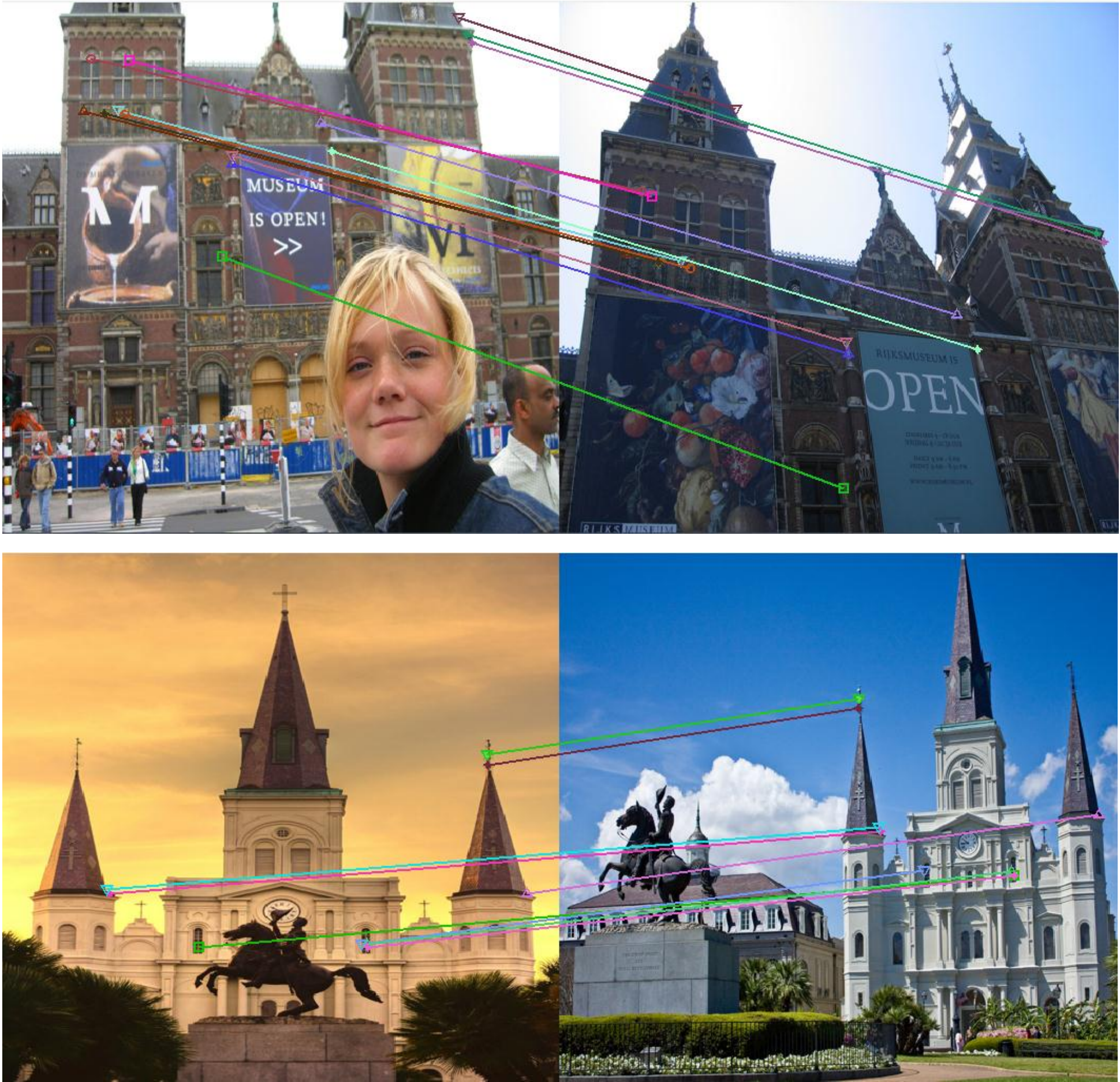}}
}
\caption{(a) When two images of the same scene are captured under considerably different conditions, many irrelevant pairs of local parts are chosen as closest pairs in the local feature space, which may lead to undesirable results of comparison.
(b) In our RDE space, matching pairs are successfully chosen as closest pairs.}
\label{fig:example_matching}
\end{figure}

However, this assumption does not hold when there are significant changes in environmental conditions (e.g., viewpoint, illumination, noise, and resolution) between two images.
For the same local part, varying environment conditions can yield varying local image patches, leading to {\em matching descriptors far apart} in the feature space.
On the other hand, for different local parts, their image patches can look similar to each other in some environmental conditions, leading to {\em non-matching descriptors close together}.
Fig. \ref{fig:example_relfar_irrnear} shows some examples:
in each triplet, the first two image patches belong to the same local part but captured under different environment conditions, while the third patch belongs to a different part but looks similar to the second one, resulting that the SIFT descriptors between non-matching local parts are closer than those between matching parts.
Consequently, comparing two images using their local descriptors cannot be done correctly when their are significant differences in environmental conditions between the images. Fig. \ref{fig:example_matching}(a) shows the cases.

In this paper, we address this problem by learning more robust representations for local image patches where matching parts are more similar together than non-matching parts even under widely varying environmental conditions.

\section{Proposed Method}

In {\em descriptor learning} \cite{HuaG2007iccv,PhilbinJ2010eccv}, a projection is obtained from training pairs of matching and non-matching descriptors in order to map given local descriptors (e.g., SIFT) to a new feature space where matching descriptors are closer to each other and non-matching descriptors are farther from each other.
Traditional techniques for supervised dimensionality reduction, including linear discriminant analysis (LDA) and local Fisher discriminant analysis (LFDA) \cite{SugiyamaM2007jmlr}, can be applied to descriptor learning after a slight modification. For example, linear discriminant embedding (LDE) \cite{HuaG2007iccv} is come from LDA with a simple modification for handling pairwise training data.

We propose a regularized learning framework in order to further emphasize (1) matching, but far apart pairs and (2) non-matching, but look-alike pairs, under wide environmental conditions.
First, we divide given training pairs of local descriptors into four subsets, {\em Relevant-Near (Rel-Near), Relevant-Far (Rel-Far), Irrelevant-Near (Irr-Near)}, and {\em Irrelevant-Far (Irr-Far)}.
For example, the ``Irr-Near'' subset consists of {\em irrelevant (i.e., non-matching), but near pairs}. We define an irrelevant pair $(\bx_i,\bx_j)$ as ``near'' if $\bx_i$ is one of the $k$ nearest descriptors\footnote{In our experiments, setting $1 \leq k \leq 10$ achieved a reasonable performance improvement.} among all non-matching descriptors of $\bx_j$ or vice versa. Similarly, a relevant pair $(\bx_i,\bx_j)$ is called ``near'' if $\bx_i$ is one of $k$ nearest descriptors among all matching descriptors of $\bx_j$. All the other pairs belong to ``Irr-Far'' or ``Rel-Far''.
Then we seek a linear projection $\bT$ that maximizes the following regularized ratio:
\begin{equation}
\label{eq:proposed_criterion}
J(\bT) = \frac{\beta_{IN} \sum_{(i,j) \in \calP_{IN}} d_{ij}(\bT) + \beta_{IF} \sum_{(i,j) \in \calP_{IF}} d_{ij}(\bT)}{\beta_{RN} \sum_{(i,j) \in \calP_{RN}} d_{ij}(\bT) + \beta_{RF} \sum_{(i,j) \in \calP_{RF}} d_{ij}(\bT)},
\end{equation}
where $d_{ij}(\bT)$ denotes the squared distance $|| \bT(\bx_i - \bx_j) ||^2$ between two local descriptors $\bx_i$ and $\bx_j$ in the projected space, and $\calP_{RN}, \calP_{RF}, \calP_{IN}, \calP_{IF}$ denote the subsets of Rel-Near, Rel-Far, Irr-Near, and Irr-Far, respectively.
Four regularization constants $\beta_{RN}, \beta_{RF}, \beta_{IN}, \beta_{IF}$ control the importance of each subset.
\begin{itemize}
\item In LDE, all pairs are equally important, i.e., $\beta_{RN} = \beta_{RF} = \beta_{IN} = \beta_{IF} = 1$.
\item In LFDA , ``near'' pairs are more important, i.e., $\beta_{RN} \gg \beta_{RF}$ and $\beta_{IN} \gg \beta_{IF}$.
\item In our method, we propose to emphasize {\em Rel-Far} (matching but far apart) and {\em Irr-Near} (non-matching but close) pairs, i.e., $\beta_{RN} \ll \beta_{RF}$ and $\beta_{IN} \gg \beta_{IF}$.
\end{itemize}

\begin{figure}[t]
\centerline{
\subfigure[]{\includegraphics[width=4cm]{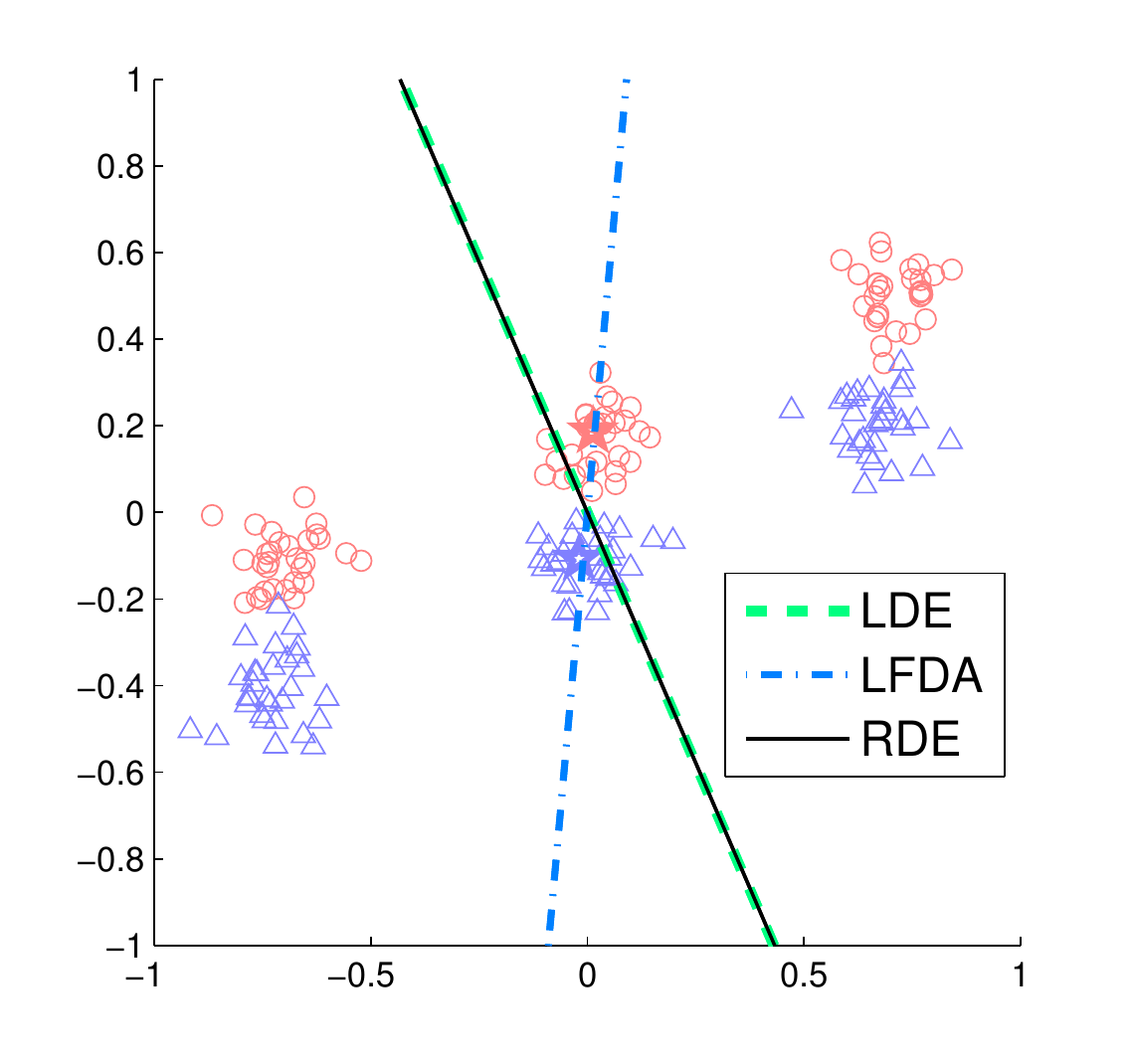}}
\hfil
\subfigure[]{\includegraphics[width=4cm]{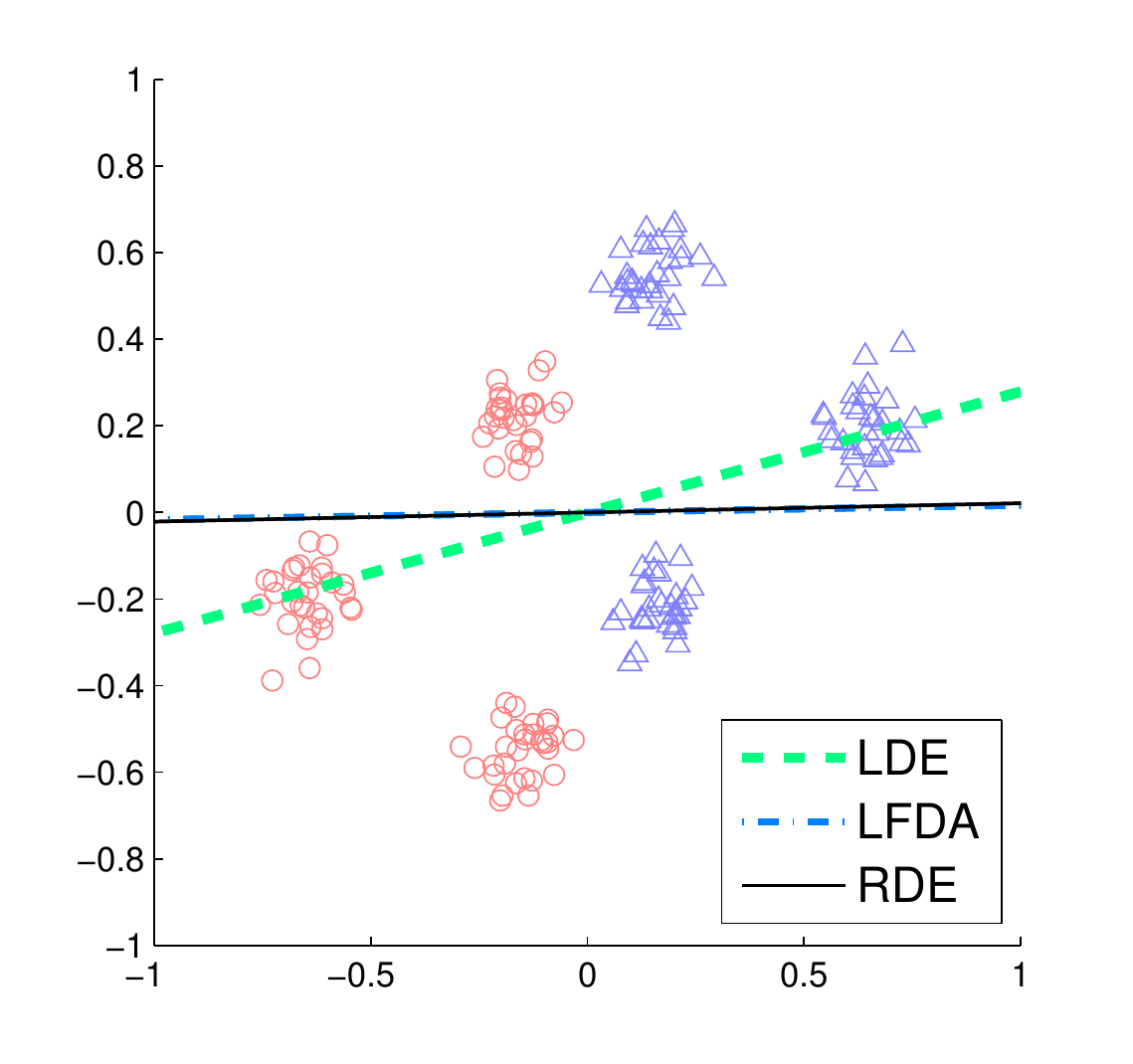}}
}
\caption{Toy examples of projections learned by LDE, LFDA, and our RDE.}
\label{fig:criteria_examples}
\end{figure}

\begin{figure}[t]
\centerline{
\subfigure[SIFT feature space]{\includegraphics[width=5cm]{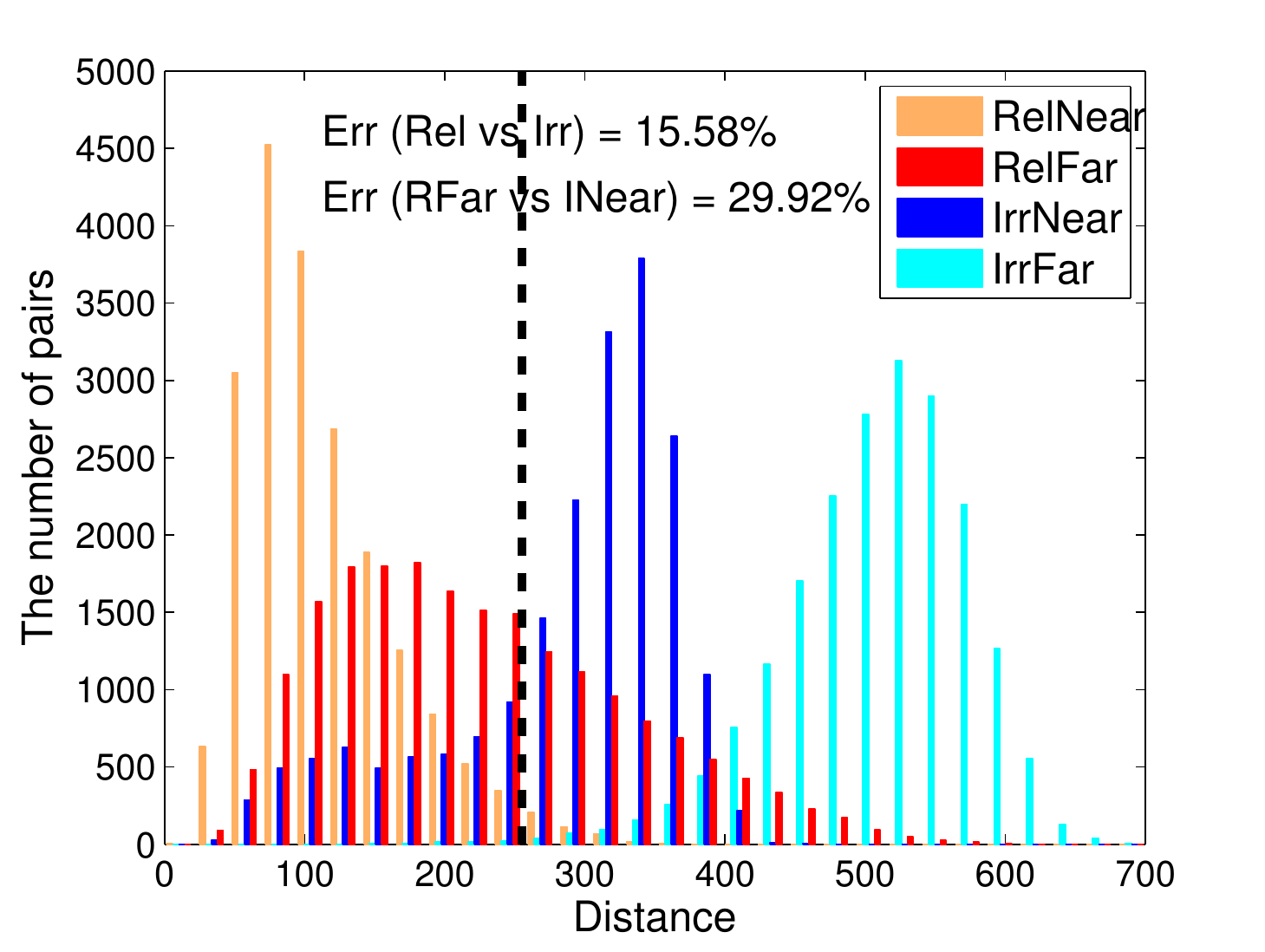}}
\hfil
\subfigure[LDE feature space]{\includegraphics[width=5cm]{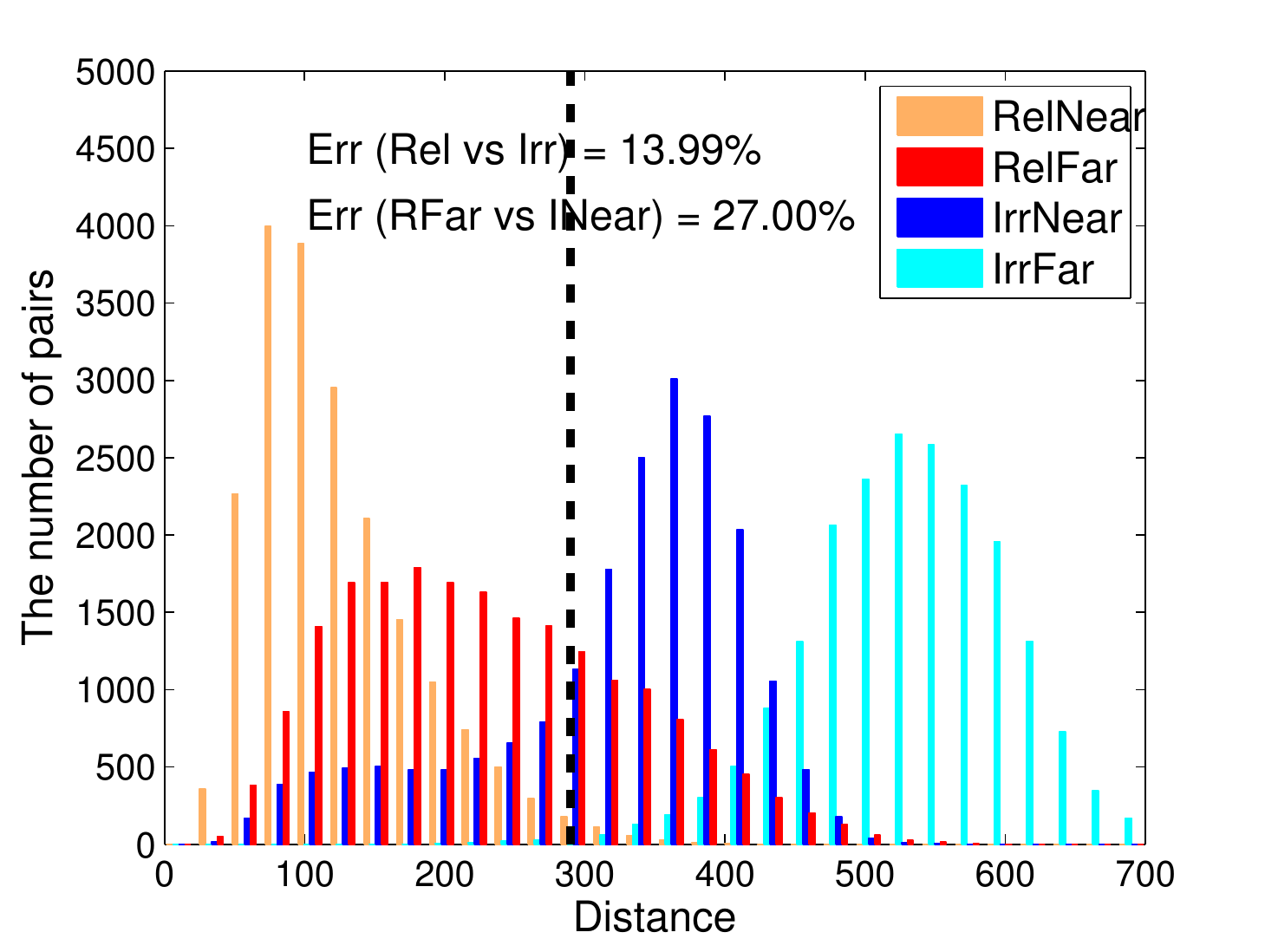}}
\hfil
\subfigure[Our RDE feature space]{\includegraphics[width=5cm]{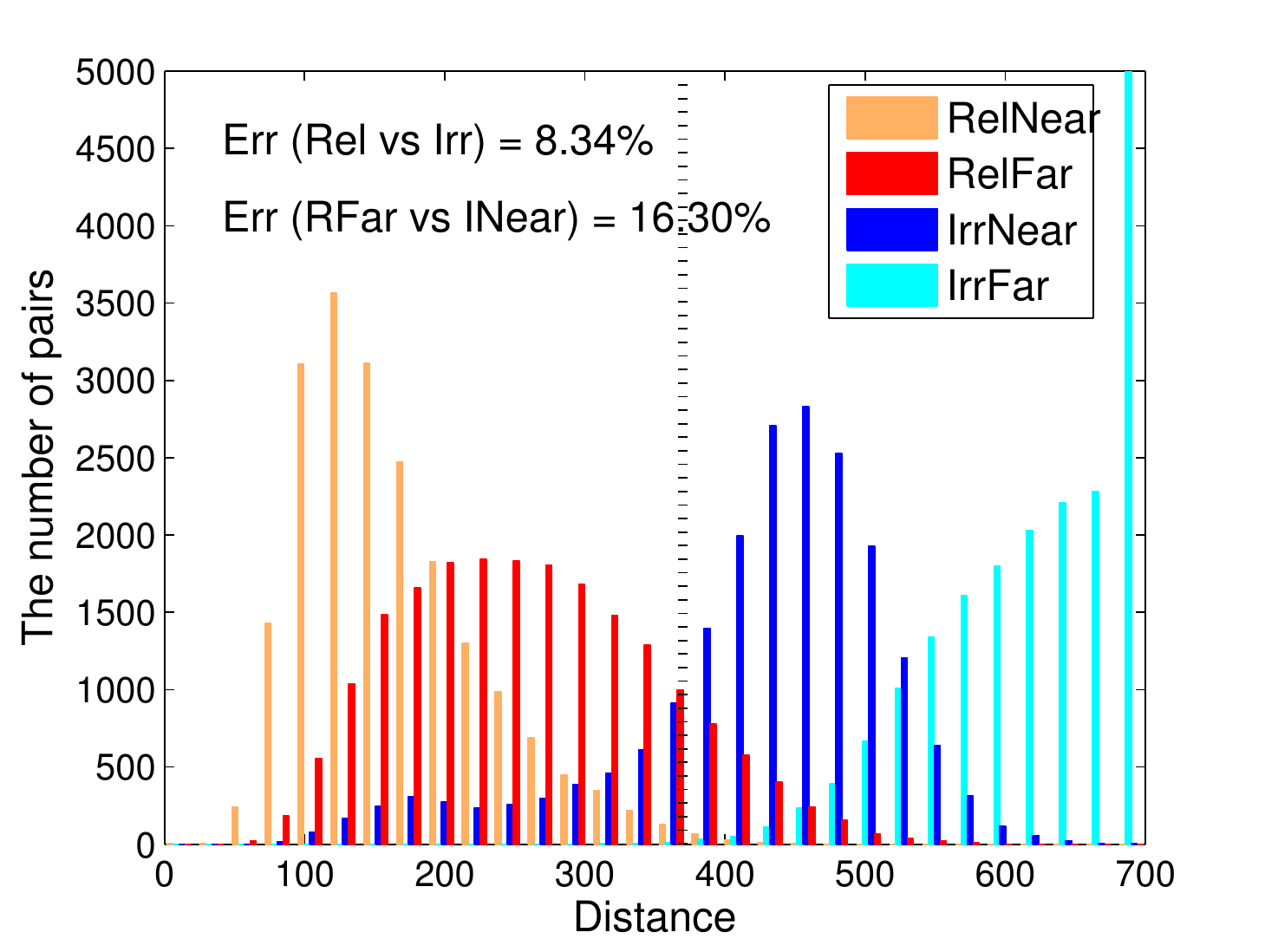}}
}
\caption{Distribution of Euclidean distance in a given feature space for each subset of pairs.
{\em Err (Rel vs Irr)} measures the proportion of overlapping region between \{Rel-Near, Rel-Far\} and \{Irr-Near, Irr-Far\}, while {\em Err (RFar vs INear)} measures the overlap between Rel-Far and Irr-Near.
In our RDE space, non-matching pairs are well distinguished from matching pairs.}
\label{fig:distance_distribution_sift}
\end{figure}

Fig. \ref{fig:criteria_examples} shows when and why our method can better distinguish Irr-Near pairs from Rel-Far pairs.
In Fig. \ref{fig:criteria_examples}(a), the global intra-class distribution forms a diagonal, while each local cluster has no meaningful direction of scattering. Since LFDA focuses on ``near'' pairs, it cannot capture the true intra-class scatter well, leading to the undesirable projection.
In Fig. \ref{fig:criteria_examples}(b), LDE obtains a projection that maximizes the inter-class variance, but the shape of the class boundary cannot be considered well, leading to an overlap between two classes. In this case, focusing more on Irr-Near pairs (i.e., the pairs of opposite clusters near the class boundary) can preserve the separability of classes.

Fig. \ref{fig:distance_distribution_sift} shows the distance distribution of local descriptors, where 20,000 pairs of each subset are randomly chosen from 500,000 local patches of Flickr images.
As shown in Fig. \ref{fig:distance_distribution_sift}(a), Rel-Near and Irr-Far pairs are already well separated in the SIFT space, but Rel-Far and Irr-Near pairs are not distinguished well ($\sim$30\% overlapped) and many Rel-Far pairs lie farther than Irr-Near pairs.
Learning by LDE can achieve only a marginal improvement (Fig. \ref{fig:distance_distribution_sift}(b)). By contrast, our RDE achieves a significant improvement in the separability between matching and non-matching pairs, especially two challenging subsets, Rel-Far and Irr-Near (Fig. \ref{fig:distance_distribution_sift}(c)).
Fig. \ref{fig:example_relfar_irrnear} and \ref{fig:example_matching} also show the superiority of our method over the existing work.

{\small
\bibliographystyle{IEEEtranS}
\bibliography{sjc}
}

\end{document}